\documentclass[11pt]{article}

\usepackage[utf8]{inputenc}
\usepackage[T1]{fontenc}
\usepackage[english]{babel}
\usepackage{lmodern}

\usepackage[margin=1in]{geometry}
\usepackage{graphicx}
\usepackage{amsmath}
\usepackage{booktabs}
\usepackage{array}
\usepackage{caption}
\usepackage{authblk}
\usepackage[super,comma,sort&compress]{natbib}
\usepackage[hidelinks]{hyperref}

\graphicspath{{figures/}}

\newcommand{\doi}[1]{doi:\,\href{https://doi.org/#1}{#1}}

\usepackage{fancyhdr}
\title{\bfseries Routine Blood Tests Outperform CRP for Distinguishing
Bacterial From Viral Infection in Children}

\author[1]{Mihaela Demireva}
\author[1]{Zhecho Mitev}
\author[1]{Djuna Chinareva-Klimentova}
\author[1]{Svetoslav Ivanov}
\author[1]{Georgi Nalbantov}
\author[2]{Dimitar Mitev}
\affil[1]{Vector Labs, Sofia, Bulgaria}
\affil[2]{Zdraveto Hospital, Sofia, Bulgaria}

\date{}

\begin{document}

\maketitle
\thispagestyle{fancy}

\begin{abstract}
\noindent
\textbf{Background:} Acute infectious diseases are among the leading causes of
medical consultations and hospitalizations in children worldwide. These
infections are predominantly caused by viruses or bacteria, yet differentiating
between the two remains a common clinical challenge. As a result, pediatricians
often default to the safer option of prescribing antibiotics contributing to the
growing problem of antimicrobial resistance. Complete Blood Count (CBC) and
C-reactive protein (CRP) tests are widely available and represent a standardized
source of information reflecting the host immune response to infection. While
the usage of blood tests to determine an infection is more often explored among
adults, little information is available on the predictive capabilities of these
tests among children. The objective is to assess the additional predictive value
of CBC towards determining the current infection.

\medskip
\noindent
\textbf{Methods:} This retrospective study used data from 906 pediatric patients
aged 2--14 years who were evaluated for suspected infection at a pediatric
hospital in Bulgaria between 2022 and 2026, focusing on the post-covid period.
These patients were tested positive either for viral or bacterial infection.
Inclusion criteria further required availability of CBC results and CRP level
measurements. These laboratory parameters as well as age were used as input
features for several supervised classification models. The commonly used method
of using CRP alone to differentiate between viral and bacterial infection was
compared to logistic regression and Gradient Boosting methods. Model performance
was evaluated using Area Under the Curve (AUC), sensitivity and specificity.

\medskip
\noindent
\textbf{Results:} The best performing model is XGBoost, which included all
features, achieving out-of-sample performance of AUC = 81.7\% and sensitivity =
70.8\%, specificity = 79.2\%. On the other hand, the XGBoost model without CRP
performs slightly worse with 1 percent less in AUC and specificity, but similar
in sensitivity. All trained models outperform a CRP-based only decision-rule
model (later mentioned as CRP baseline model) in terms of AUC.

\medskip
\noindent
\textbf{Conclusions:} Both logistic regression (linear) and XGBoost (non-linear)
models better distinguish between viral and bacterial infection compared to the
CRP baseline model. We suggest that the decision to prescribe antibiotics should
be based on a number of factors, including but not limited to CBC, some of which
are not currently incorporated into routine practice. Factors such as white
blood count (WBC), lymphocyte count (LYM) and monocyte percentage (MON\%) appear
to be more informative than the value of CRP for the distinction according to
our best XGBoost model.
\end{abstract}

\section{Introduction}

Despite significant advances in medicine and technology, distinguishing between
viral and bacterial infections remains a difficult task. Symptoms such as fever,
muscle pain, and generalized weakness may be caused by either type, while the
required treatments differ substantially. Early and accurate determination of
the infection type is crucial for prescribing appropriate therapy---bacterial
infections usually require antibiotic treatment, whereas antibiotics can be
ineffective against viral infections.\cite{tanday2016}

Antibiotics overprescription is a well-known problem, still researched actively.
Wrong usage of antibiotics can lead to development of bacterial resistance to
antibiotics.\cite{llor2014}

The doctors' practice tends to be different among the different countries. A
pediatric study from 5 European countries\cite{sanz2005} examines the
correctness of antibiotics prescription. The researchers found that the
incorrect prescription of antibiotics reached a maximum of 64.7\% in Bratislava,
Slovakia. Moreover, according to the study, in western countries more doctors
prescribed wrong antibiotics as a first line medication in case of Tonsillitis
or Otitis media. A more recent study conducted in the United
States\cite{harris2016} indicates that at least 30\% of prescribed antibiotics
are unnecessary, despite the implementation of programs aimed at reducing
inappropriate antibiotic use.\cite{fiore2017}

In case of symptoms, hinting bacterial or viral infection, doctors might use
external factors to decide whether to prescribe or not antibiotics. Zaykova et
al.\ (2024)\cite{zaykova2024} found evidence for different approaches among
different types of physicians. Counterintuitively, decisions also varied by
weekday versus weekend/holiday periods.

A common practice among doctors is to use CRP (C-reactive Protein) tests as an
indicator of the existence of bacterial infection and therefore as a decision
rule whether to prescribe antibiotics or not. Medical papers and laboratories
define low CRP level to indicate viral infection, while high value --
bacterial. The range of 10--40~mg/L is considered critical, since the infection
can be either bacterial or viral.\cite{guncar2024} However, there are 2 main
reasons why one can not use CRP as the only identification for the type of
infection. First, when CRP is below the critical range, bacterial infection is
still possible especially if the infection started less than 48 hours before the
test.\cite{markanday2015} Second, adenovirus, influenza and SARS-CoV-2 could
cause CRP to rise above 40~mg/L, therefore higher levels could not exclude the
presence of a virus.

Various biomarkers have been investigated for their ability to differentiate
infection types in adults. Multiple authors suggest that combining multiple
biomarkers can improve the accuracy of the prediction
outcome.\cite{gille2012,oved2015} It is also believed that some key blood
indicators vary in terms of gender and age.\cite{adeli2015,doucoure2024}
Therefore such factors must also be taken into account for a more accurate
classification. More recently, machine learning approaches have demonstrated
improved diagnostic accuracy by integrating multiple blood parameters.
Gun\v{c}ar et al.\ (2024)\cite{guncar2024} developed a model using 16 routine
blood test results along with CRP, achieving 82.2\% accuracy in distinguishing
bacterial from viral infections in adults---outperforming CRP-based decision
rules alone.

In pediatric populations, distinguishing infection etiology presents additional
challenges due to the relatively immature immune system, different epidemiology
of infections, and difficulties in specimen collection compared to
adults.\cite{tsao2020} Several studies have focused on febrile children,
particularly infants and preschool-aged children.\cite{lee2022}

Other studies examining infection differentiation in children have focused on
novel biomarkers requiring specialized assays,\cite{srugo2017,vanhouten2017}
while the combined diagnostic potential of routinely available tests---complete
blood count and CRP---remains underexplored in pediatric populations. In our
research, we address this gap by evaluating the utility of CBC parameters
combined with CRP for distinguishing viral from bacterial infections in children
aged 2--14 years. Our goal is to evaluate what is the contribution of CRP as a
feature in such a classification model and how much better a classification
model performs compared to the simple decision rule using CRP only.

\section{Materials and methods}

\subsection{Patient Population}

Between 2022 and 2026, a total of 5347 pediatric patients between 2 and 14 years
with suspected infection have been tested for either viral, bacterial or both
infections in hospital Zdraveto. We applied the following eligibility criteria:
proven bacteria or viral infection; available CBC and CRP tests. As a result, in
3617 patients an infection was identified. Since our focus is on differentiating
viral or bacterial infections, only observations with confirmed infections from
any of the both kinds of tests were kept. Five patients have been positive for
both viral and bacterial infection, thus due to ambiguity they have been
removed. Out of the remaining 3612 patients only a cohort of 956 patients have
no missing values for all blood tests (CBC, CRP). 50 patients, which are the
most recent, are taken out and used later for held-out validation. Therefore our
final cohort consists of 906 patients. Out of these, 424 are with bacterial
infection and 482 are with viral infection.

Descriptive statistics of all features are given in Table~\ref{tab:descriptive}.

\begin{table}[htbp]
\centering
\caption{Descriptive statistics of model parameters in the whole dataset and
split by output (Virus versus Bacteria)}
\label{tab:descriptive}
\small
\begin{tabular}{l c c c}
\toprule
 & \textbf{All ($n=906$)} & \textbf{Virus ($n=482$)} & \textbf{Bacteria ($n=424$)} \\
\midrule
age   & $5.05 \pm 2.78$     & $4.9 \pm 2.86$      & $5.23 \pm 2.69$ \\
CRP   & $16.66 \pm 27.97$   & $11.73 \pm 18.4$    & $22.27 \pm 35.06$ \\
LYM   & $2.25 \pm 1.24$     & $1.94 \pm 1.14$     & $2.6 \pm 1.25$ \\
MON   & $0.74 \pm 0.37$     & $0.65 \pm 0.34$     & $0.83 \pm 0.38$ \\
NEU   & $6.36 \pm 4.98$     & $4.62 \pm 2.96$     & $8.33 \pm 5.98$ \\
RBC   & $4.88 \pm 0.4$      & $4.89 \pm 0.41$     & $4.87 \pm 0.38$ \\
HGB   & $131.08 \pm 10.09$  & $130.9 \pm 10.09$   & $131.28 \pm 10.09$ \\
HCT   & $0.4 \pm 0.03$      & $0.4 \pm 0.03$      & $0.4 \pm 0.03$ \\
MCV   & $81.47 \pm 4.58$    & $81.47 \pm 4.6$     & $81.48 \pm 4.56$ \\
MCHC  & $330.79 \pm 8.73$   & $329.93 \pm 8.72$   & $331.78 \pm 8.66$ \\
RDW   & $13.1 \pm 1.57$     & $13.17 \pm 1.39$    & $13.01 \pm 1.76$ \\
WBC   & $9.51 \pm 5.36$     & $7.32 \pm 3.45$     & $12.0 \pm 6.03$ \\
MPV   & $8.96 \pm 0.89$     & $9.01 \pm 0.9$      & $8.9 \pm 0.87$ \\
LYM\% & $27.69 \pm 14.94$   & $28.91 \pm 14.38$   & $26.3 \pm 15.46$ \\
MON\% & $8.35 \pm 3.31$     & $9.18 \pm 3.42$     & $7.42 \pm 2.92$ \\
NEU\% & $61.46 \pm 16.95$   & $59.88 \pm 16.19$   & $63.27 \pm 17.62$ \\
MCH   & $26.96 \pm 1.68$    & $26.88 \pm 1.67$    & $27.04 \pm 1.7$ \\
PLT   & $265.65 \pm 79.07$  & $242.55 \pm 76.38$  & $291.92 \pm 73.78$ \\
PCT   & $0.26 \pm 0.23$     & $0.23 \pm 0.16$     & $0.29 \pm 0.29$ \\
PDW   & $15.84 \pm 0.41$    & $15.84 \pm 0.42$    & $15.83 \pm 0.4$ \\
\bottomrule
\end{tabular}

\vspace{0.6em}
\parbox{0.9\textwidth}{\small Values are mean $\pm$ SD.
Abbreviations: $n$, number; SD, standard deviation.}
\end{table}

\subsection{Data processing and labelling}

Several viral and bacterial diagnostic tests were performed. Viral testing
included assays for SARS-CoV-2, influenza A and B, respiratory syncytial virus
(RSV), and adenovirus. Due to the fact that test results were in text form, data
preprocessing was required in order to define the final virus label as positive
or negative.

Some of the patients have been tested for bacteria through quick tests for
streptococci and/or nasal or throat secretions. For the latter, additional data
preprocessing has been made since results indicate the presence and the type of
bacteria as well as its intensity in text form. We exclude the ambiguous cases
and leave only those with no bacterial finding or with proven presence of such.

Our outcome variable is defined as follows: ``1'': presence of bacterial
infection and ``0'': presence of viral infection.

\subsection{Statistical analysis}

Using the dataset described above, we formulated the prediction task as a binary
classification problem. Two main well-known supervised machine learning
approaches were considered and compared -- logistic regression and
XGBoost.\cite{chen2016} For each of which a hyperparameter search was performed
in order to optimize our chosen performance criteria -- Area under the curve
(AUC). For the logistic regression all features have been min-max scaled. Along
with the four mentioned models, a univariate logistic regression analysis was
also performed.

Apart from age and CRP, the standard parameters from a CBC have been used as
input variables:

\begin{quote}
\textbf{LYM} -- Lymphocytes (absolute count),
\textbf{LYM\%} -- Lymphocyte percentage,
\textbf{MON} -- Monocytes (absolute count),
\textbf{MON\%} -- Monocyte percentage,
\textbf{NEU} -- Neutrophils (absolute count),
\textbf{NEU\%} -- Neutrophil percentage,
\textbf{RBC} -- Red Blood Cell count,
\textbf{HGB} -- Hemoglobin,
\textbf{HCT} -- Hematocrit,
\textbf{MCV} -- Mean Corpuscular Volume,
\textbf{MCH} -- Mean Corpuscular Hemoglobin,
\textbf{MCHC} -- Mean Corpuscular Hemoglobin Concentration,
\textbf{RDW} -- Red Cell Distribution Width,
\textbf{WBC} -- White Blood Cell count,
\textbf{PLT} -- Platelet count,
\textbf{MPV} -- Mean Platelet Volume,
\textbf{PCT} -- Plateletcrit,
\textbf{PDW} -- Platelet Distribution Width.
\end{quote}

Both logistic regression and XGBoost models have hyperparameters which have to
be selected for a final model. For logistic regression we use L2-norm
penalization whose shrinkage parameter is the only hyperparameter that must be
tuned. For XGBoost the following parameters were tuned: maximum tree depth
(\textit{max\_depth}), number of boosting rounds (\textit{n\_estimators}),
feature subsampling ratio (\textit{colsample\_bytree}), minimum child weight
(\textit{min\_child\_weight}), minimum loss reduction required to make a split
(\textit{gamma}),fraction of observations used for each tree (\textit{subsample}) and L2 regularization (\textit{reg\_lambda}).

A standard procedure for hyperparameter selection is the 5-fold cross-validation
with stratification selection. For each hyperparameter combination we obtain
cross-validation out-of-sample AUC. The same 5-fold split was used for all
models to ensure fair comparison. Final models were built with the
hyperparameters which produced the highest cross-validation AUC values.

The chosen models were also compared with a CRP Baseline Model using only CRP as
a predictor. Similar to other authors\cite{guncar2024} we chose an optimal
cutoff value for the CRP Baseline Model based on the highest accuracy from the
whole dataset. We note that this is in-sample performance, therefore
overoptimistic. In our case the cut-off is 22~mg/L. According to this rule all
cases below 22~mg/L were considered viral infection, and all equal to 22 and
above -- bacterial.

Additionally, SHAP (SHapley Additive exPlanations) provides a unified method for
feature importance and is widely used for model
interpretability.\cite{lundberg2017} We computed SHAP values on the final model
trained on the full dataset.

\section{Results}

First, 12 variables were found to be statistically significant (all except for
RBC, HGB, HCT, MCV, RDW, MPV, MCH and PDW) in an univariate non-penalized
logistic regression analysis (Table~\ref{tab:univariate}, Supplementary
Materials). Given the large sample size, the $p$-values should be interpreted
with some caution, as they may be inflated in significance.

In the multivariate case, our results were obtained using 5-fold cross
validation with stratification on the earlier described dataset with 906
observations. We compare 2 machine learning algorithms, each with 2 models --
one with CRP as a feature and one without CRP using AUC, sensitivity and
specificity as performance metrics. To assess the statistical performance of the
best models and the CRP baseline model 95\% confidence intervals were estimated
by bootstrapping the out-of-fold predictions of the final selected models using
1000 bootstrap resamples.\cite{davison1997,hastie2009}

\begin{figure}[htbp]
\centering
\includegraphics[width=0.70\textwidth]{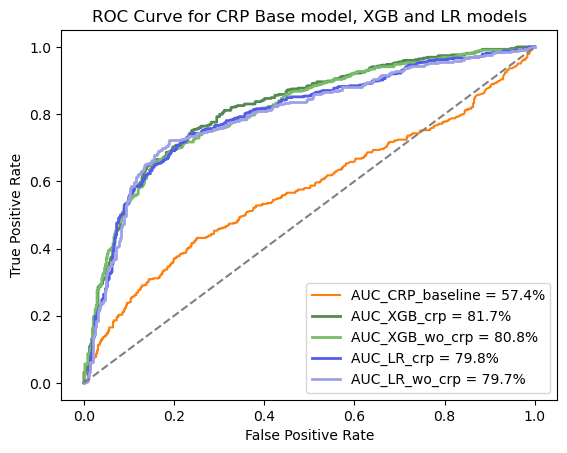}
\caption{Receiver operating characteristic (ROC) curves on all trained models
using out-of-fold predictions from 5-fold cross validation with stratification
and CRP Baseline model on the whole dataset. AUC\_CRP\_baseline -- Area Under
the Curve for baseline model with CRP alone; AUC\_XGB\_crp -- Area Under the
Curve for XGBoost model with CRP as variable; AUC\_XGB\_wo\_crp -- Area Under
the Curve for XGBoost model without CRP as variable; AUC\_LR\_crp -- Area Under
the Curve for logistic regression model with CRP as variable; AUC\_LR\_wo\_crp
-- Area Under the Curve for logistic regression model without CRP as variable.}
\label{fig:roc}
\end{figure}

Our best performing model is XGBoost with CRP as an input feature (see
Table~\ref{tab:performance}, Figure~\ref{fig:roc}). It achieves an area under
the curve (AUC) of 81.7\%, with a sensitivity of 70.8\% and a specificity of
79.2\%. The XGBoost model trained without CRP as a feature performs slightly
worse in terms of AUC (AUC -- 80.8\%, sensitivity -- 71\%, specificity -- 78\%).
As shown in Table~\ref{tab:performance} for both cases XGBoost performed better
than logistic regression in sensitivity and in AUC. However, the specificity of
both logistic regression models is higher compared to the XGBoost models on the
account of sensitivity of ${\sim}67\%$.

\begin{figure}[htbp]
\centering
\includegraphics[width=0.70\textwidth]{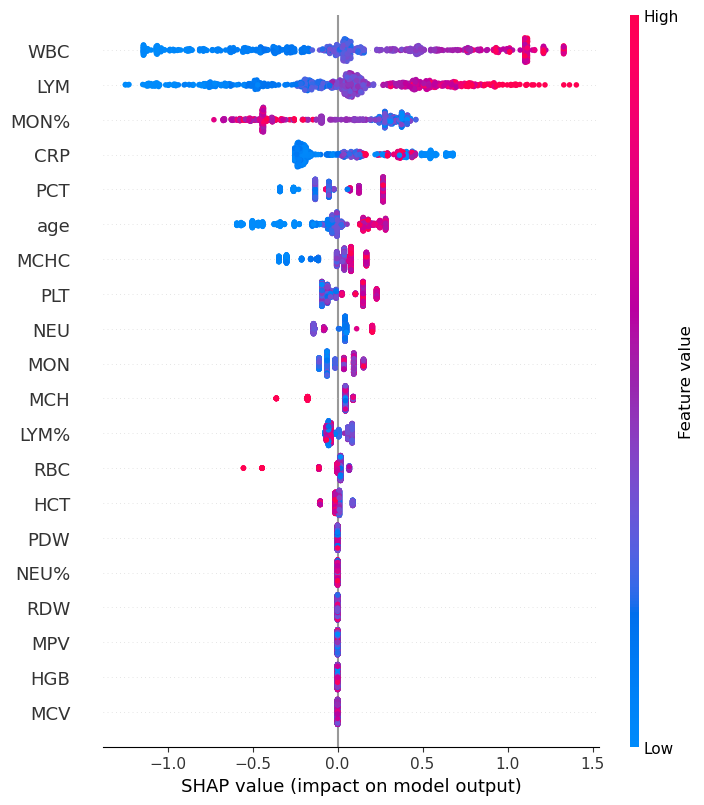}
\caption{SHAP values for the whole dataset on the final model trained. Each dot
represents an observation and the color and its intensity represent the effect
of that feature on the dot. Effects to the right of 0.0 are positive, and to the
left -- negative. WBC, LYM, MON\%, CRP, PCT and age have the highest impact, and
PDW, NEU\%, RDW, MPV, HGB and MCV -- the lowest.}
\label{fig:shap}
\end{figure}

As shown in Figure~\ref{fig:shap}, the SHAP analysis identified white blood
count (WBC), lymphocyte count (LYM) and monocyte percentage (MON\%) as the most
influential features of the best mode. Higher WBC and LYM values generally
increased the probability of bacterial infection, while lower WBC and LYM values
shifted predictions toward viral infection. Other laboratory parameters had
comparatively smaller effects, with most SHAP values clustered near zero (e.g.\
MCV, HGB, MPW, RDW, NEU\%, PDW).

Considering AUC, all our models perform substantially better than the CRP
Baseline model at all thresholds, which only has an AUC of 57.4\%
(Figure~\ref{fig:roc}).

Finally, we compared our XGBoost model with CRP as a feature to a CRP Baseline
model with a cut-off 22~mg/L. The CRP Baseline model achieved a sensitivity of
30.9\% and a specificity of 85.6\%, being outperformed significantly by all four
trained models.

To validate our findings, we have tested all models on a held-out set consisting
of 50 patients (16 viral, 34 bacterial). The pattern of results remained
consistent (see Table~\ref{tab:heldout}, Supplementary Materials). Even on
unseen data the trained models add value compared to the CRP Baseline model,
which has an AUC barely above the random guess (52.8\%). Given the small size
and pronounced class imbalance of this held-out set (68\% bacterial vs.\ 46.8\%
in training), class-specific metrics such as specificity and sensitivity should
be interpreted with caution.

\begin{table}[htbp]
\centering
\caption{Model performance with 95\% confidence intervals using bootstrap
resampling of the out-of-fold predictions (1000 iterations).}
\label{tab:performance}
\small
\begin{tabular}{l c c c}
\toprule
Model & AUC & Sensitivity (Recall 1) & Specificity (Recall 0) \\
\midrule
XGBoost w/ CRP      & 81.7\% [78.8\%--84.3\%] & 70.8\% [66.1\%--75.2\%] & 79.2\% [75.6\%--82.8\%] \\
XGBoost wo/ CRP     & 80.8\% [77.8\%--83.6\%] & 71.0\% [66.4\%--75.3\%] & 78.0\% [74.4\%--81.5\%] \\
LR w/ CRP           & 79.8\% [76.8\%--82.6\%] & 66.5\% [61.9\%--71.2\%] & 82.1\% [78.7\%--85.6\%] \\
LR wo/ CRP          & 79.7\% [76.7\%--82.5\%] & 67.9\% [63.3\%--72.5\%] & 84.2\% [81.2\%--87.3\%] \\
CRP Baseline model  & 57.4\% [53.4\%--61.3\%] & 30.9\% [26.4\%--35.3\%] & 85.6\% [82.4\%--88.5\%] \\
\bottomrule
\end{tabular}

\vspace{0.6em}
\parbox{0.95\textwidth}{\small Abbreviations: w/ CRP -- model including CRP as a
feature; wo/ CRP -- model excluding CRP as a feature; LR -- Logistic Regression
model. XGBoost w/ CRP is the best performing model overall (AUC = 81.7\% and
sensitivity = 70.8\%, specificity = 79.2\%).}
\end{table}

\clearpage

\section{Discussion}

In this research, we have employed a statistical approach capable of
differentiating between viral and bacterial infections in children. Our results
suggest that the combination of multiple blood indicators outperforms the
predictions produced by the CRP baseline.

According to our best performing model, the most important input features are
WBC, LYM and MON\%. All these features were shown to be statistically
significant in the univariate logistic regression analysis as well
(Table~\ref{tab:univariate}, Supplementary Materials). Other researchers show
that a virus such as SARS-CoV-2 is leading to a decrease in the absolute
lymphocyte count.\cite{huang2020} Our findings indicate that lymphopenia may
serve as a broader indicator of viral infection beyond COVID-19 as the period we
are covering is post-pandemic (after 2022).

Elevated WBC levels are commonly associated with bacterial infections,
reflecting the immune system's response through increased production of these
cells to fight the invading pathogens.\cite{riley2015} Our research confirms
this observation throughout our study population. Our model places WBC as the
most important indicator for the differentiation of virus vs bacteria. This also
aligns with other statistical studies, confirming WBC to be an important
indicator for discovering bacterial infections.\cite{gille2012}

It is important to note that none of these biomarkers alone are sufficient to
make a reliable classification of infection. Our results indicate that a
sophisticated combination of multiple factors is required to achieve optimal
results.

C-reactive protein is widely regarded as an important indicator of bacterial
infection.\cite{yo2012} However, our study suggests that CRP alone has limited
discriminative ability for differentiating viral from bacterial infections. This
is supported by the histograms of patients grouped by infection type (viral vs.\
bacterial), which show substantial overlap and similar means between the two
classes (Figure~\ref{fig:crphist}, Supplementary Material). Despite selecting
the optimal threshold of 22~mg/L for the CRP baseline model it produces a
specificity slightly above our best performing model (85.6\% vs 79.2\%), and a
sensitivity much lower than XGBoost (30.9\% vs 70.8\%).

Even when combined with complete blood count (CBC) parameters, CRP contributed
only marginally to model performance. This was reflected in both machine
learning algorithms (XGBoost and logistic regression), where inclusion of CRP
resulted in only a modest improvement in AUC (1 percentage point for XGBoost and
0.2 percentage points for logistic regression).

Other authors developing machine learning-based models have used similar models
for adults\cite{guncar2024} and very young children ($<$6 years
old).\cite{lee2022} They have utilized similar CBC parameters and CRP as input
features. However, they note that CRP is a very good indicator of the type of
infection, while in our cohort of children, CRP is not crucial for the
distinction. This discrepancy may be attributable to differences in age
distribution, timing of biomarker assessment, or clinical setting, as CRP
kinetics and baseline inflammatory responses can vary across
populations.\cite{doucoure2024,sproston2018} This is a limitation of our study
since we have not analyzed the differences in the age distribution due to the
lack of enough data per age group.

The findings of this study may support specialists in making more informed
decisions when evaluating children with suspected infection. Our method of
combining routine blood tests, such as complete blood count and CRP, provides a
way to make distinction between viral and bacterial infections. This may help
pediatricians reduce the antibiotic prescriptions or recognise early the need
for one, which could lead to more appropriate treatment overall.

While this research showed significant results, it has several limitations.
First, all models were trained on data collected from a single hospital, which
may limit the generalizability of our findings. Second, the time elapsed between
illness onset of the patients and the timing of laboratory examinations were not
available. The absence of this information prevents the model from better
understanding of the stages of the illness.

Future work could involve developing a multicentric study including observations
from other hospitals, capturing other relationships between CBC parameters, CRP
measurement and infection type. Furthermore, we could compare our model output
with the decisions that pediatricians make based on the same variables and check
if the model has a better prediction accuracy compared to practitioners.

\section{Conclusion}

A classification model was developed for distinguishing between viral and
bacterial infections in children. The models include complete blood count and
CRP as input features and were all found to outperform the CRP Baseline model.
Moreover, some CBC input features add more contribution to the overall
predictive performance of the model compared to CRP. WBC, LYM and MON\% appear
most informative to our model. A larger amount of data from various hospitals is
required to validate our findings and improve the model performance.
Nonetheless, the results of our study show that such a model has the potential
to be used as a tool facilitating general practitioners and pediatricians' in
prescription decisions.


\clearpage

\appendix
\renewcommand{\thetable}{S\arabic{table}}
\renewcommand{\thefigure}{S\arabic{figure}}
\setcounter{table}{0}
\setcounter{figure}{0}

\section*{Supplementary Material}

\begin{table}[htbp]
\centering
\caption{Univariate logistic regression analysis}
\label{tab:univariate}
\small
\begin{tabular}{l r r r}
\toprule
Variable & Coefficient & S.E. & $p$-value \\
\midrule
age   & $0.041823$  & $0.023973$ & $.081$   \\
CRP   & $0.016445$  & $0.003156$ & $<.001$  \\
LYM   & $0.515430$  & $0.067262$ & $<.001$  \\
MON   & $1.411930$  & $0.201374$ & $<.001$  \\
NEU   & $0.206319$  & $0.020504$ & $<.001$  \\
RBC   & $-0.122709$ & $0.168858$ & $.467$   \\
HGB   & $0.003723$  & $0.006609$ & $.573$   \\
HCT   & $-1.258714$ & $2.195184$ & $.566$   \\
MCV   & $0.000582$  & $0.014560$ & $.968$   \\
MCHC  & $0.024618$  & $0.007768$ & $.002$   \\
RDW   & $-0.062658$ & $0.044088$ & $.155$   \\
WBC   & $0.237452$  & $0.020533$ & $<.001$  \\
MPV   & $0.137576$  & $0.075931$ & $.070$   \\
LYM\% & $-0.011787$ & $0.004519$ & $.009$   \\
MON\% & $-0.177906$ & $0.023293$ & $<.001$  \\
NEU\% & $0.011988$  & $0.004008$ & $.003$   \\
MCH   & $0.056400$  & $0.040218$ & $.161$   \\
PLT   & $0.009552$  & $0.001069$ & $<.001$  \\
PCT   & $4.542359$  & $1.087153$ & $<.001$  \\
PDW   & $-0.085368$ & $0.163949$ & $.603$   \\
\bottomrule
\end{tabular}

\vspace{0.6em}
\parbox{0.8\textwidth}{\small Abbreviations: S.E., standard error.}
\end{table}

\begin{table}[htbp]
\centering
\caption{Held-out set results ($n=50$, Virus = 16, Bacteria = 34)}
\label{tab:heldout}
\small
\begin{tabular}{l c c c c}
\toprule
Model & Validation AUC & Accuracy & Sensitivity & Specificity \\
\midrule
CRP baseline model & 52.8\% & 44\% & 24\% & 88\% \\
XGBoost w/ CRP     & 74.1\% & 68\% & 68\% & 69\% \\
XGBoost wo/ CRP    & 74.8\% & 66\% & 65\% & 69\% \\
LR w/ CRP          & 74.8\% & 70\% & 74\% & 62\% \\
LR wo/ CRP         & 71.5\% & 70\% & 74\% & 62\% \\
\bottomrule
\end{tabular}
\end{table}

\begin{figure}[htbp]
\centering
\includegraphics[width=0.70\textwidth]{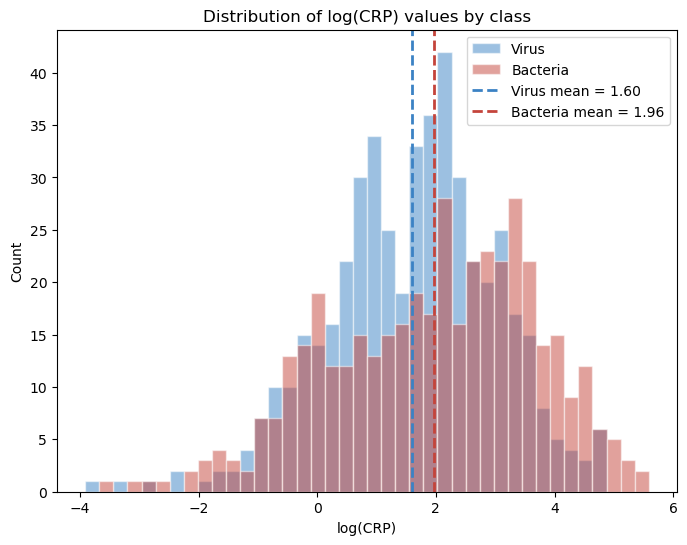}
\caption{Distribution of log values of CRP grouped by type of infection --
viral or bacterial.}
\label{fig:crphist}
\end{figure}


\begin{thebibliography}{99}
\setlength{\itemsep}{2pt}

\bibitem{tanday2016}
Tanday S. Resisting the use of antibiotics for viral infections.
\textit{Lancet Respir Med.} 2016;4(3):179.

\bibitem{llor2014}
Llor C, Bjerrum L. Antimicrobial resistance: risk associated with antibiotic
overuse and initiatives to reduce the problem.
\textit{Ther Adv Drug Saf.} 2014;5(6):229--241.

\bibitem{sanz2005}
Sanz Emilio J, Hernandez Miguel A, Ratchina S, et al. Prescribers' indications
for drugs in childhood: a survey of five European countries (Spain, France,
Bulgaria, Slovakia and Russia). \textit{Acta Paediatr.} 2005;94(12):1784--1790.
\doi{10.1111/j.1651-2227.2005.tb01854.x}

\bibitem{harris2016}
Harris AM, Hicks LA, Qaseem A; High Value Care Task Force of the American
College of Physicians and for the Centers for Disease Control and Prevention.
Appropriate antibiotic use for acute respiratory tract infection in adults:
advice for high-value care from the American College of Physicians and the
Centers for Disease Control and Prevention.
\textit{Ann Intern Med.} 2016;164(6):425--434. \doi{10.7326/M15-1840}

\bibitem{fiore2017}
Fiore DC, Fettic LP, Wright SD, Ferrara BR. Antibiotic overprescribing: still a
major concern. \textit{J Fam Pract.} 2017;66(12):730--736.

\bibitem{zaykova2024}
Zaykova K, Nikolova SP, Pancheva R, Serbezova A. Antibiotic prescribing
practices to children among in- and outpatient physicians in Bulgaria.
\textit{Acta Medica Bulgaria.} 2024;51(4). \doi{10.2478/amb-2024-0075}

\bibitem{guncar2024}
Gun\v{c}ar G, Kukar M, Smole T, et al. Differentiating viral and bacterial
infections: a machine learning model based on routine blood test values.
\textit{Heliyon.} 2024;10(8):e29372. \doi{10.1016/j.heliyon.2024.e29372}

\bibitem{markanday2015}
Markanday A. Acute phase reactants in infections: evidence-based review and a
guide for clinicians. \textit{Open Forum Infect Dis.} 2015;2(3):ofv098.
\doi{10.1093/ofid/ofv098}

\bibitem{gille2012}
Gille-Johnson P, Hansson KE, G{\aa}rdlund B. Clinical and laboratory variables
identifying bacterial infection and bacteraemia in the emergency department.
\textit{Scand J Infect Dis.} 2012;44(10):745--752.
\doi{10.3109/00365548.2012.689846}

\bibitem{oved2015}
Oved K, Cohen A, Boico O, et al. A novel host-proteome signature for
distinguishing between acute bacterial and viral infections.
\textit{PLoS One.} 2015;10(3):e0120012. \doi{10.1371/journal.pone.0120012}

\bibitem{adeli2015}
Adeli K, Raizman JE, Chen Y, et al. Complex biological profile of hematologic
markers across pediatric, adult, and geriatric ages: establishment of robust
pediatric and adult reference intervals on the basis of the Canadian Health
Measures Survey. \textit{Clin Chem.} 2015;61(8):1075--1086.
\doi{10.1373/clinchem.2015.240531}

\bibitem{doucoure2024}
Doucoure MB, Wright JK, Criss AH, et al. Normal clinical laboratory ranges by
age and sex: relationship with complete blood count and chemistry parameters.
\textit{Clin Lab Sci.} 2024;37(2).

\bibitem{tsao2020}
Tsao YT, Tsai YH, Liao WT, et al. Differential markers of bacterial and viral
infections in children for point-of-care testing.
\textit{Trends Mol Med.} 2020;26(12):1118--1132.
\doi{10.1016/j.molmed.2020.09.004}

\bibitem{lee2022}
Lee B, Chung HJ, Kang HM, Kim DK, Kwak YH. Development and validation of machine
learning-driven prediction model for serious bacterial infection among febrile
children in emergency departments. \textit{PLoS One.} 2022;17(3):e0265500.
\doi{10.1371/journal.pone.0265500}

\bibitem{srugo2017}
Srugo I, Klein A, Stein M, et al. Validation of a novel assay to distinguish
bacterial and viral infections. \textit{Pediatrics.} 2017;140(4):e20163453.

\bibitem{vanhouten2017}
van Houten CB, de Groot JAH, Klein A, et al. A host-protein based assay to
differentiate between bacterial and viral infections in preschool children
(OPPORTUNITY): a double-blind, multicentre, validation study.
\textit{Lancet Infect Dis.} 2017;17(4):431--440.

\bibitem{chen2016}
Chen T, Guestrin C. XGBoost: a scalable tree boosting system. In:
\textit{Proceedings of the 22nd ACM SIGKDD International Conference on Knowledge
Discovery and Data Mining (KDD '16).} New York, NY: Association for Computing
Machinery; 2016:785--794. \doi{10.1145/2939672.2939785}

\bibitem{lundberg2017}
Lundberg SM, Lee SI. A unified approach to interpreting model predictions. In:
\textit{Advances in Neural Information Processing Systems 30 (NIPS 2017).}
Red Hook, NY: Curran Associates; 2017:4765--4774.

\bibitem{davison1997}
Davison AC, Hinkley DV. \textit{Bootstrap Methods and Their Application.}
Cambridge: Cambridge University Press; 1997.

\bibitem{hastie2009}
Hastie T, Tibshirani R, Friedman J. \textit{The Elements of Statistical
Learning: Data Mining, Inference, and Prediction.} 2nd ed. Springer; 2009.

\bibitem{huang2020}
Huang C, Wang Y, Li X, et al. Clinical features of patients infected with 2019
novel coronavirus in Wuhan, China.
\textit{The Lancet.} 2020;395(10223):497--506.

\bibitem{riley2015}
Riley LK, Rupert J. Evaluation of patients with leukocytosis.
\textit{Am Fam Physician.} 2015;92(11):1004--1011.

\bibitem{yo2012}
Yo CH, Hsieh PS, Lee SH, et al. Comparison of the test characteristics of
procalcitonin to C-reactive protein and leukocytosis for the detection of
serious bacterial infections in children presenting with fever without source: a
systematic review and meta-analysis.
\textit{Ann Emerg Med.} 2012;60(5):591--600.

\bibitem{sproston2018}
Sproston NR, Ashworth JJ. Role of C-reactive protein at sites of inflammation
and infection. \textit{Front Immunol.} 2018;9:754.

\end{thebibliography}
\end{document}